\documentclass{article}

\usepackage{microtype}
\usepackage{graphicx}
\usepackage{subfigure}
\usepackage{booktabs} 

\usepackage{wrapfig}
\usepackage{colortbl}
\usepackage{color}
\definecolor{light}{rgb}{0.3, 0.3, 0.3}
\def\light#1{{\color{light}#1}}

\usepackage{hyperref}

\usepackage[accepted]{icml2025}

\usepackage{amsmath}
\usepackage{amssymb}
\usepackage{mathtools}
\usepackage{amsthm}
\usepackage{multirow}

\usepackage[capitalize,noabbrev]{cleveref}

\theoremstyle{plain}

\theoremstyle{definition}

\theoremstyle{remark}

\usepackage[textsize=tiny]{todonotes}

\icmltitlerunning{ToxBench: A Binding Affinity Prediction Benchmark with AB-FEP-Calculated Labels for Human Estrogen Receptor Alpha}

\begin{document}

\twocolumn[
\icmltitle{ToxBench: A Binding Affinity Prediction Benchmark with AB-FEP-Calculated Labels for Human Estrogen Receptor Alpha}



\icmlsetsymbol{equal}{*}

\begin{icmlauthorlist}
\icmlauthor{Meng Liu}{nvda,equal}
\icmlauthor{Karl Leswing}{sdgr,equal}
\icmlauthor{Simon K. S. Chu}{nvda}
\icmlauthor{Farhad Ramezanghorbani}{nvda}
\icmlauthor{Griffin Young}{sdgr}
\icmlauthor{Gabriel Marques}{sdgr}
\icmlauthor{Prerna Das}{sdgr}
\icmlauthor{Anjali Panikar}{sdgr}
\icmlauthor{Esther Jamir}{sdgr}
\icmlauthor{Mohammed Sulaiman Shamsudeen}{sdgr}
\icmlauthor{
K. Shawn Watts}{sdgr}
\icmlauthor{Ananya Sen}{sdgr}
\icmlauthor{Hari Priya Devannagari}{sdgr}
\icmlauthor{Edward B. Miller}{sdgr}
\icmlauthor{Muyun Lihan}{sdgr}
\icmlauthor{Howook Hwang}{sdgr}
\icmlauthor{Janet Paulsen}{nvda}
\icmlauthor{Xin Yu}{nvda}
\icmlauthor{Kyle Gion}{nvda}
\icmlauthor{Timur Rvachov}{nvda}
\icmlauthor{Emine Kucukbenli}{nvda}
\icmlauthor{Saee Gopal Paliwal}{nvda}
\end{icmlauthorlist}

\icmlaffiliation{nvda}{NVIDIA}
\icmlaffiliation{sdgr}{Schr{\"o}dinger}

\icmlcorrespondingauthor{Meng Liu}{menliu@nvidia.com}
\icmlcorrespondingauthor{Karl Leswing}{karl.leswing@schrodinger.com}

\icmlkeywords{Machine Learning, ICML}

\vskip 0.3in
]



\printAffiliationsAndNotice{\icmlEqualContribution} 

\begin{abstract}
Protein-ligand binding affinity prediction is essential for drug discovery and toxicity assessment. While machine learning (ML) promises fast and accurate predictions, its progress is constrained by the availability of reliable data. In contrast, physics-based methods such as absolute binding free energy perturbation (AB-FEP) deliver high accuracy but are computationally prohibitive for high-throughput applications. To bridge this gap, we introduce ToxBench, the first large-scale AB-FEP dataset designed for ML development and focused on a single pharmaceutically critical target, Human Estrogen Receptor Alpha (ER$\alpha$). ToxBench contains 8,770 ER$\alpha$-ligand complex structures with binding free energies computed via AB-FEP with a subset validated against experimental affinities at 1.75 \textit{kcal/mol} RMSE, along with non-overlapping ligand splits to assess model generalizability. Using ToxBench, we further benchmark state-of-the-art ML methods, and notably, our proposed DualBind model, which employs a dual-loss framework to effectively learn the binding energy function. The benchmark results demonstrate the superior performance of DualBind and the potential of ML to approximate AB-FEP at a fraction of the computational cost. 
\end{abstract}

\section{Introduction}
\label{sec:intro}

\begin{figure}[t]
	\centering
\includegraphics[width=0.48\textwidth]{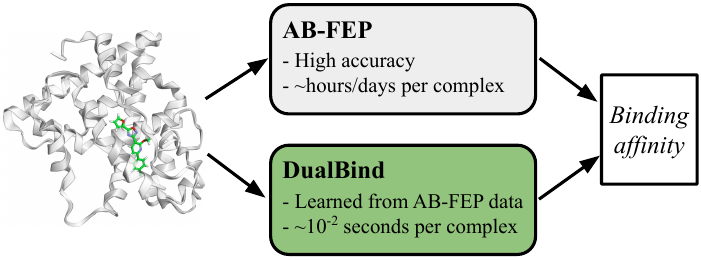}
	\caption{\textbf{An illustration of the ToxBench task.} ML models, such as DualBind, are trained on high-fidelity AB-FEP data to predict protein-ligand binding affinities several orders of magnitude faster than original AB-FEP calculations.}
	\label{fig:toxbench_task}
\end{figure}


Fast and accurate prediction of protein-ligand binding affinity is fundamental for modern drug discovery. The ability to reliably estimate how tightly a small molecule binds to its target protein enables the prioritization of drug candidates and the early identification of off-target interactions that could lead to adverse effects~\citep{meng2011molecular,pagadala2017software,chen2018rise,yang2019concepts,vamathevan2019applications,beroza2022chemical,sadybekov2023computational}. Machine learning (ML) models offer the potential for rapid and accurate predictions, but their success critically depends on the availability of reliable datasets for training and evaluation. Unfortunately, obtaining large-scale and high-quality experimental affinity data remains a significant bottleneck due to the resource-intensive nature of experimental affinity measurements. 


Existing affinity datasets have attempted to address this challenge but are shown to suffer from inherent biases that constrain model generalizability. Particularly, ML models are typically trained on the PDBBind dataset~\citep{wang2004pdbbind,liu2017forging} and  evaluated on the CASF-2016 benchmark~\citep{su2018comparative}. However, recent studies indicate that models developed using these datasets tend to learn dataset-specific biases rather than truly capturing the underlying protein-ligand interactions~\citep{volkov2022frustration,li2024leak,durant2025robustly}. Notably, as detailed in Section~\ref{sec:related_work}, models trained solely on ligand or protein features, without modeling protein-ligand interactions, have achieved unexpectedly competitive performance on CASF-2016~\citep{yang2020predicting,volkov2022frustration}. This underscores the limitations of current benchmarks and highlights the urgent need for a more robust testbed that reflects real-world challenges.

In contrast to data-driven ML approaches, absolute protein-ligand binding free energy calculation through free energy perturbation (AB-FEP) has emerged as a highly accurate method.~\citep{jorgensen1988efficient,aldeghi2016accurate,Chen2023}. By employing rigorous all-atom molecular dynamics (MD) simulations in explicit solvent, AB-FEP produces binding free energy calculations with accuracy comparable to experimental assays, which typically have uncertainties around 1.0 \textit{kcal/mol} \citep{Ross2023maximal, Davis2011}. Notably, the AB-FEP implementation in Schr{\"o}dinger's FEP+~\citep{schrodinger2021fep} achieves a root mean square error (RMSE) of $\sim$1.1 \textit{kcal/mol} against experimental affinities in validation studies~\citep{Chen2023}. However, due to extensive MD simulations, the AB-FEP calculation is computationally intensive and can often take hours to days to complete for a single complex system, thus limiting its practical use in high-throughput applications.


In response to the above challenges, in this paper, we introduce ToxBench, the first AB-FEP dataset centered on Human Estrogen Receptor Alpha (ER$\alpha$), a pharmaceutically critical target. ER$\alpha$ is central to endocrine signaling and a key target for both therapeutic development and toxicity assessment, as its modulation is linked to various adverse outcomes, such as reproductive disorders and hormone-dependent cancers~\citep{lee2013molecular,la2020consensus,miziak2023estrogen}. In particular, ToxBench comprises 8,770 ER$\alpha$-ligand complexes with binding free energies computed via AB-FEP and partially validated against available experimental measurements. By incorporating non-overlapping ligand splits and concentrating on a single, pharmaceutically critical target, ToxBench closely aligns with real-world structure-based virtual screening scenarios, where extensive ligand libraries are assessed against one critical target. Therefore, ToxBench serves as a realistic testbed for developing and evaluating ML models for binding affinity prediction.

Using ToxBench, we conduct a comprehensive benchmarking study in which we train and evaluate representative ML models for binding affinity prediction. In addition, we propose a novel model, DualBind, which integrates supervised mean squared error (MSE) loss with unsupervised denoising score matching (DSM) loss to effectively learn the binding energy function. Our experimental results highlight the superior performance of DualBind and demonstrate that carefully designed ML models have the potential to approximate AB-FEP at a fraction of the computational cost, as illustrated in Figure~\ref{fig:toxbench_task}.

Our contributions can be summarized as follows:
\begin{itemize}
    \item \textbf{ToxBench Dataset}: We present ToxBench, the first large-scale AB-FEP dataset focused on the critical ER$\alpha$ target. ToxBench is designed as a realistic testbed for development of binding affinity prediction models.
    \item \textbf{DualBind Model}: We propose DualBind, a novel deep learning framework that synergistically combines MSE and DSM losses, leading to superior performance.
    \item \textbf{Benchmarking Study}: We conduct benchmarking experiments for state-of-the-art ML models on ToxBench, establishing robust baselines to guide future research in binding affinity prediction.
\end{itemize}

The dataset is available via~\url{https://huggingface.co/datasets/karlleswing/toxbench} and the DualBind implementation can be found at~\url{https://github.com/NVIDIA-Digital-Bio/dualbind}.


\section{Related Work}
\label{sec:related_work}

\textbf{Binding Affinity Prediction}. Binding affinity prediction methods generally fall into three categories: scoring functions, physics-based approaches, and ML models. Conventional scoring functions~\citep{bohm1994development,head1996validate,eldridge1997empirical,bohm1998prediction,gohlke2000predicting,wang2002further}, usually based on empirical terms or knowledge-based analyses, offer rapid estimates but often lack accuracy, particularly when applied outside their training domains. In contrast, physics-based approaches, such as MM-PBSA~\citep{kollman2000calculating,homeyer2012free}, MM-GBSA~\citep{still1990semianalytical,gohlke2003insights,gohlke2004converging}, and AB-FEP~\citep{jorgensen1988efficient,gilson1997statistical,boresch2003absolute}, deliver more reliable predictions but are computationally extensive due to the high cost of molecular dynamics (MD) simulations and solvent modeling. In contrast, ML methods aim to achieve both speed and accuracy. Recent advances have introduced a variety of data-driven ML models for binding affinity prediction, built on sequence-based~\citep{ozturk2018deepdta,yuan2022fusiondta}, graph-based~\citep{nguyen2021graphdta,moon2022pignet}, or 3D structure-based~\citep{jimenez2018k,jiang2021interactiongraphnet,lu2022tankbind} representations and neural networks. Our proposed DualBind model, which considers spatial information, falls into the 3D structure-based category. Given the vast literature in this field, we refer readers to surveys such as~\citet{meng2011molecular,meli2022scoring,liu2024binding} for a comprehensive review of existing binding affinity prediction methods.

\textbf{Existing Datasets and Limitations}. The effectiveness of ML models for protein-ligand binding affinity prediction heavily depends on the quality of the datasets used for training and evaluation. Specifically, PDBBind~\citep{wang2004pdbbind,liu2017forging}, which includes $\sim$20k protein-ligand complex structures with affinity measurements collected from various sources, and its 285-complex core set which is also known as CASF-2016~\citep{su2018comparative}, have become the de facto benchmarks in the field. ML models are typically trained on PDBBind and evaluated on the CASF-2016 benchmark. However, recent work by~\citet{volkov2022frustration} demonstrates that, under this setup, models using only ligand or protein features, without modeling protein-ligand interactions, perform surprisingly well and modeling interactions does not provide clear advantage. In fact,~\citet{volkov2022frustration}'s experiments reveal that a model trained solely on ligand graphs achieves a Pearson correlation coefficient ($R_p$) of 0.749 and an RMSE of 1.567, outperforming the model trained on protein-ligand interaction graphs ($R_p$ = 0.687, RMSE = 1.605). This finding has been independently confirmed by multiple studies~\citep{yang2020predicting,wang2022yuel,durant2025robustly}. Together, these studies suggest that inherent biases in existing datasets cause models to rely on dataset-specific ligand-only or protein-only signals, rather than truly learning the underlying protein-ligand interactions. As a result, such trained models often fail to generalize to novel complexes~\citep{volkov2022frustration,scantlebury2023small}.

Notably,~\citet{volkov2022frustration} also defines the density of a training set as $$D=\frac{\#\text{observed complexes}}{\#\text{proteins}\times\#\text{ligands}},$$ and reports $D<0.05$ for PDBBind, meaning that over 95\% of all possible protein–ligand pairs lack training affinity labels. Experimental studies from~\citet{volkov2022frustration} indicate that, in such sparse regimes, models often exploit ligand-only or protein-only features as shortcuts, rather than learning protein-ligand interactions, and that increasing density is an attractive path to increase model generalizability. This insight motivates ToxBench. By focusing on a single target (ER$\alpha$) and providing several thousands of AB-FEP-labeled complexes, ToxBench achieves full density for this receptor. This encourages ML models to truly learn protein-ligand interactions, thus improving their ability to predict affinities for novel ligands. We argue that such a task design aligns well with the practical target-based virtual screening scenario.

\section{ToxBench Benchmark}

\textbf{Dataset Generation}. The construction of the ToxBench dataset started with obtaining a set of ligand compounds targeting the Human Estrogen Receptor Alpha (ER$\alpha$). The compound dataset was constructed by integrating data from the ChEMBL~\citep{Zdrazil2024} database and the Diverse Unbiased Validation-Extended (DUD-E)~\citep{Mysinger2012} dataset. An initial set of 473 compounds with reported assay values was obtained from ChEMBL. From this collection, 67 experimental binding affinity values were manually curated and standardized to units of \textit{kcal/mol}. These specific values were prioritized due to their high data quality and consistent reporting as relative binding affinities. Furthermore, the binding affinity data for three specific ChEMBL compounds (CHEMBL234638, CHEMBL236086 and CHEMBL236718), initially reported in \textit{Ki}, were validated and converted to \textit{kcal/mol}.

The structural representation of ER$\alpha$ in ToxBench was derived from three Protein Data Bank (PDB)~\citep{Berman2000} entries, which were refined using the Schrödinger Protein Preparation Wizard~\citep{Sastry_2013}. To account for the conformational diversity of ER$\alpha$, PDB structure 1ERE was selected as representative of the agonist-bound conformation, while PDB structure 3ERT was selected to represent the antagonist-bound conformation. Additionally, PDB entry 1SJ0, which captures ER$\alpha$ in another distinct conformational state, was included to further diversify the protein templates.

Ligand preparation was conducted using Schrödinger LigPrep with Epik7 ~\citep{Johnston2023Epik, Schrodinger2025LigPrep}. For ChEMBL-derived ligands, a protocol involving Induced Fit Docking/Molecular Dynamics (IFD/MD)~\citep{Miller2021} was employed, followed by Absolute Binding Free Energy Perturbation (AB-FEP) ~\citep{Chen2023} simulations performed for 1 ns. For ligands sourced from the DUD-E dataset, protein-ligand poses were generated using GLIDE-WS ~\citep{Halgren2004, Murphy2016WScore}, also followed by 1 ns AB-FEP simulations. 

\begin{figure}[t]
\centering
\includegraphics[width=0.3\textwidth]{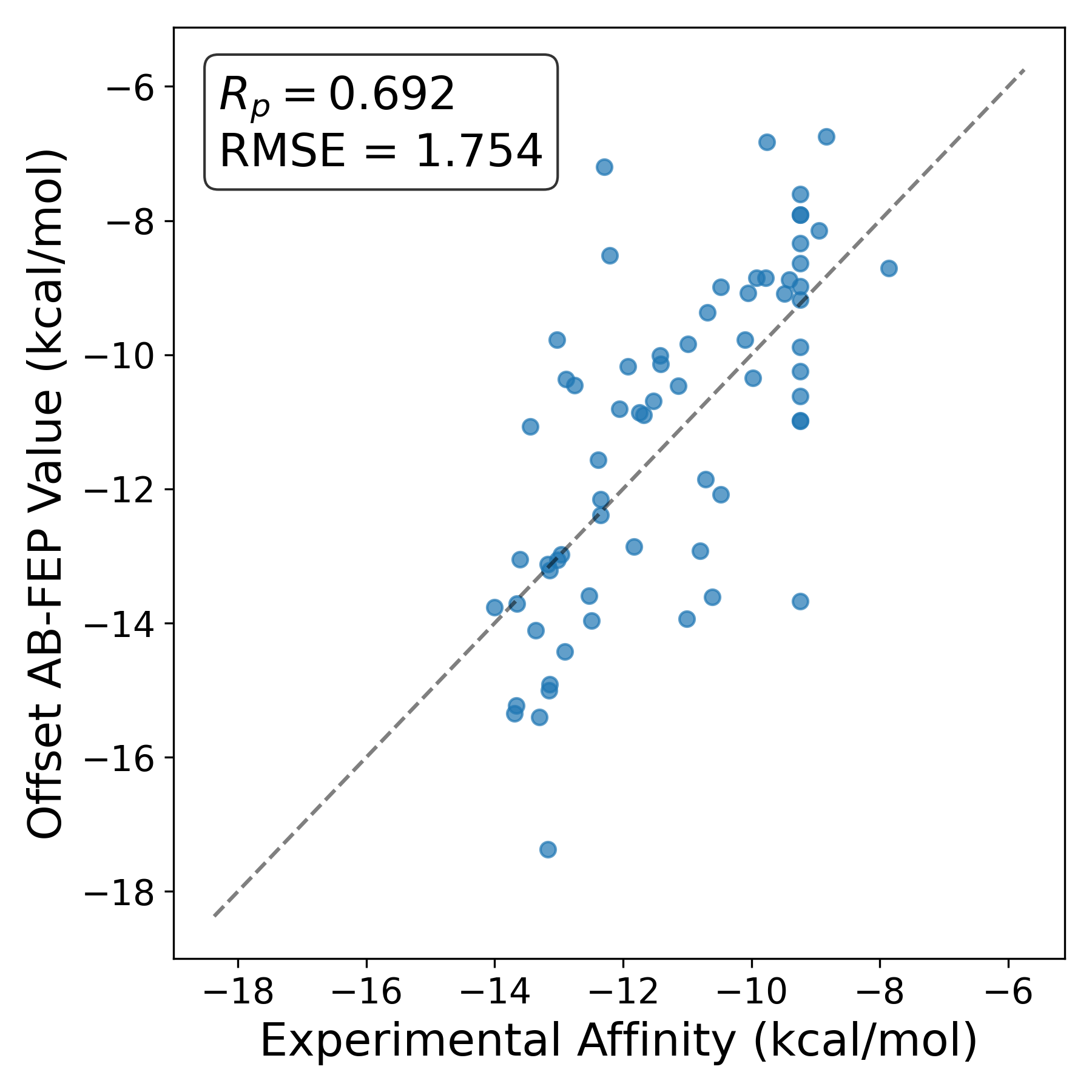}
 \caption{\textbf{AB-FEP calculations \emph{vs.} experimental affinities}. This Compares AB-FEP calculated binding affinities with the corresponding 67 curated experimental binding affinities.}
 \label{fig:ABFEP_vs_Experiment}
\end{figure}

\textbf{Agreement of Calculations with Experimental Data}. To assess the accuracy of our AB-FEP calculations, we compared them against the 67 curated experimental binding affinities. To obtain a single representative computational binding affinity towards ER$\alpha$ for each of the 67 manually verified experimental values, the ensemble of AB-FEP calculations was systematically processed. An empirical, PDB-specific structural reorganization penalty was applied to the calculated free energies: +8.69 \textit{kcal/mol} for the 1ERE structure and +6.11 \textit{kcal/mol} for the 3ERT structure. Following these adjustments, the lowest (most favorable) free energy value across all sampled conformations for each ligand was designated as its final computational binding affinity. This workflow constitutes an initial algorithmic strategy to correlate the experimentally observed binding affinities with the predictions derived from conformational sampling and free energy calculations. As illustrated in Figure~\ref{fig:ABFEP_vs_Experiment}, this measurement yielded an RMSE of 1.754 \textit{kcal/mol} and a Pearson correlation coefficient ($R_p$) of 0.692 when compared with the curated experimental data set. While the $R_p$ value might appear modest, particularly when compared to $R_p$ values from ML models evaluated on the ToxBench test set (Section~\ref{sec:experiments}), this can be primarily attributed to the significantly narrower dynamic range of binding affinities within this specific 67-compound experimental validation set. It is well-known that $R_p$ values are sensitive to the data range, with narrow ranges often leading to lower $R_p$ values for a similar level of absolute error. The RMSE of 1.754 \textit{kcal/mol}, however, serves as a direct measure of absolute error. Achieving this level of accuracy validates the practical utility of our AB-FEP calculations, as it approaches the $\sim$1.0 \textit{kcal/mol} uncertainty typically associated with experimental binding assays, thereby supporting the reliability of the ToxBench labels.

\begin{figure}[t]
\centering
\includegraphics[width=0.3\textwidth]{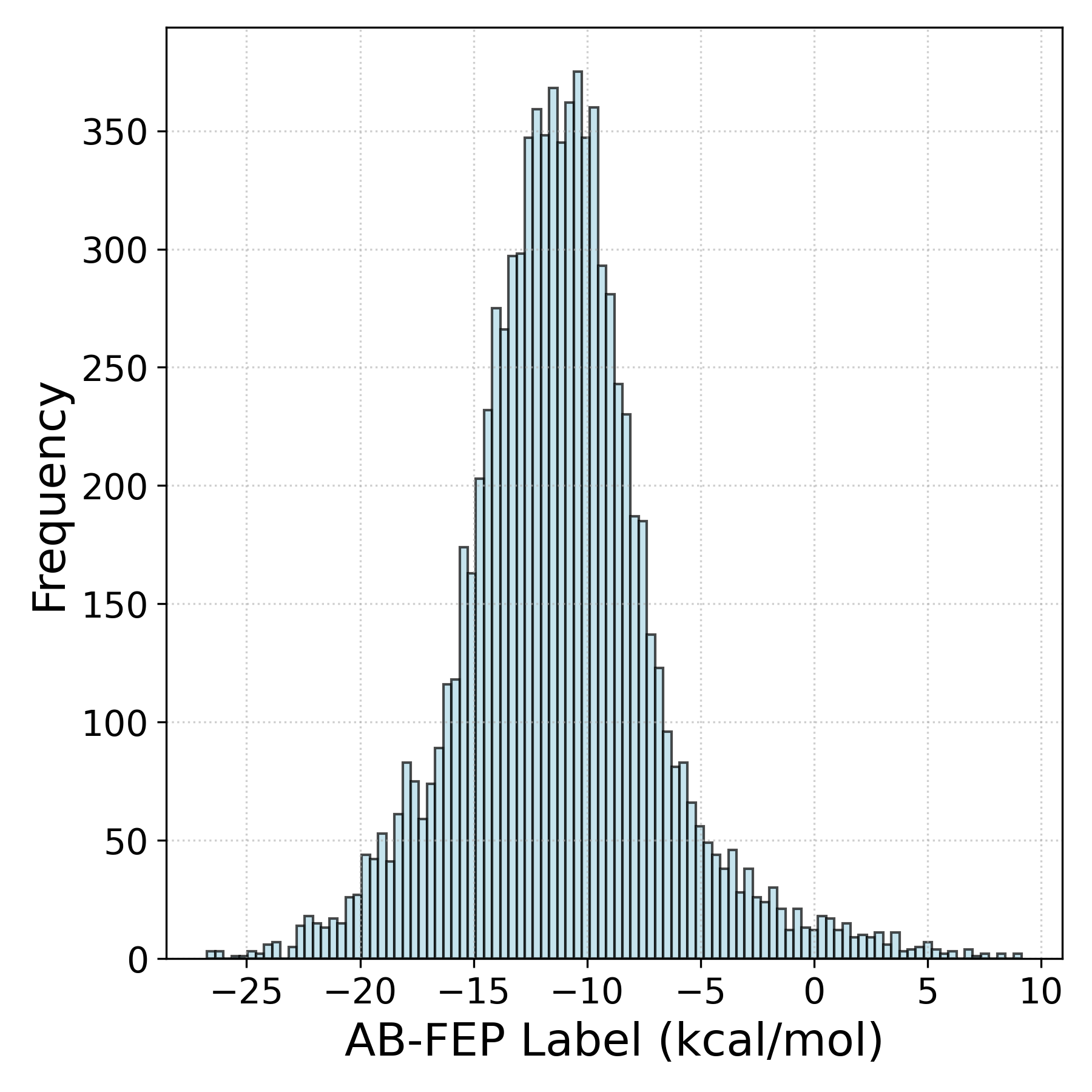}
 \caption{\textbf{Distribution of the AB-FEP calculated binding affinity labels in ToxBench}. These labels, representing binding free energies for the 8,770 ER$\alpha$-ligand complexes, are in units of \textit{kcal/mol} and span a range of approximately -26 to +9 \textit{kcal/mol}.}
 \label{fig:ABFEP_distribution}
\end{figure}

\textbf{Data Statistics}. In total, ToxBench contains 8,770 ER$\alpha$-ligand complexes covering 699 unique ligand SMILES, each annotated with an AB-FEP-calculated binding free energy. Using a 70\%/15\%/15\% random split and ensuring no SMILES overlap, we obtain 6,144 training data (covering 488 unique SMILES), 1,317 validation data (covering 102 unique SMILES), and 1,309 test data (covering 109 unique SMILES). The AB-FEP values span a range of approximately -26 to +9 \textit{kcal/mol}. The overall distribution is shown in Figure~\ref{fig:ABFEP_distribution}.

By having a single-target focus, high-fidelity AB-FEP labels, and non-overlapping ligand splits, ToxBench offers a valuable testbed for developing and assessing ML models in binding affinity prediction.


\section{DualBind Method}
\label{sec:DualBind}

Here, we further propose a new approach, namely DualBind, for binding affinity prediction. DualBind employs a novel dual-loss strategy, which combines supervised mean squared error (MSE) loss with unsupervised denoising score matching (DSM) loss to effectively learn the binding energy function.

\begin{figure*}[t]
	\centering
\includegraphics[width=0.75\textwidth]{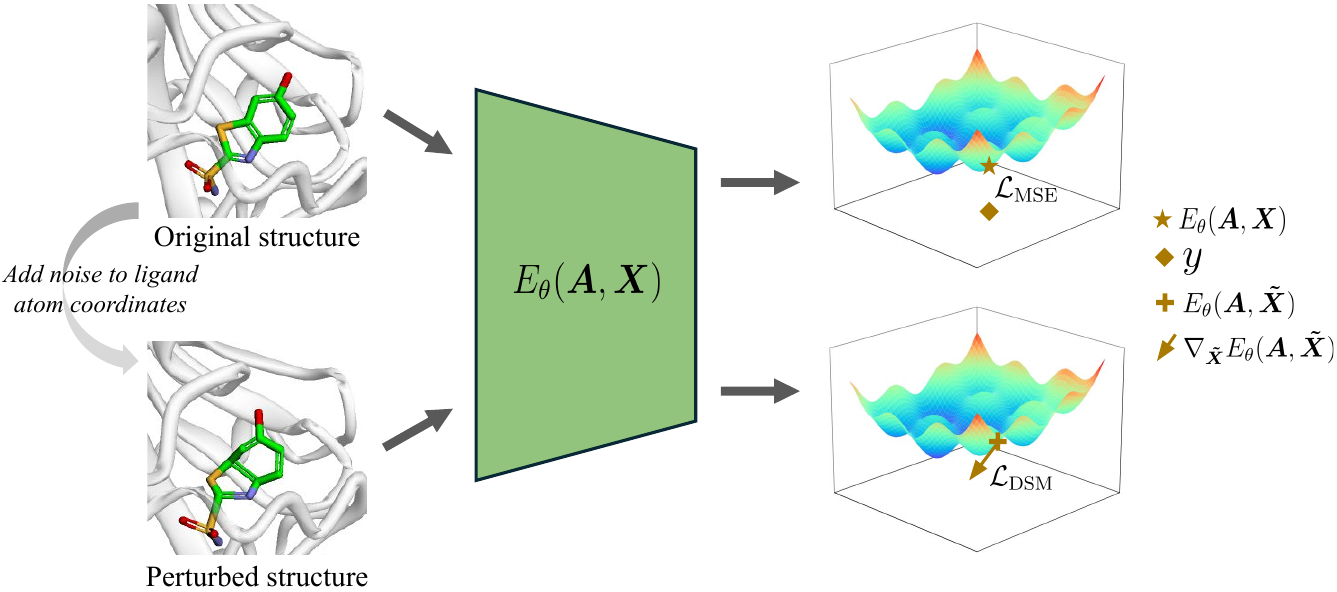}
	\caption{\textbf{An illustration of the DualBind approach.} DualBind employs a dual-loss framework that combines the MSE loss $\mathcal{L}_{\text{MSE}}$ and the DSM loss $\mathcal{L}_{\text{DSM}}$. Specifically, $\mathcal{L}_{\text{MSE}}$ anchors the predicted binding affinity of the original structure to its groundtruth label. Concurrently, $\mathcal{L}_{\text{DSM}}$ shapes the gradient of the energy function at the perturbed structure. Details are described in Section~\ref{sec:DualBind}.}
	\label{fig:DualBind}
\end{figure*}

\textbf{The Dual-Loss Framework.} We represent a protein-ligand complex structure as $C = (\boldsymbol{A}, \boldsymbol{X})$, where $\boldsymbol{A} \in \mathbb{R}^{n \times d}$ denotes the features of the $n$ atoms in the complex, including both protein and ligand atoms, and $\boldsymbol{X} \in \mathbb{R}^{n \times 3}$ represents their atomic coordinates. We define a parameterized binding affinity prediction model as $E_\theta(\boldsymbol{A}, \boldsymbol{X}): \mathbb{R}^{n \times d} \times \mathbb{R}^{n \times 3} \rightarrow \mathbb{R}$, which produces a scalar energy value for a given complex structure. $\theta$ denotes the learnable parameters in the model.

Intuitively, the dual-loss framework in DualBind uses the DSM loss $\mathcal{L}_{\text{DSM}}$ to shape the energy landscape by guiding the gradient of the energy function, while the MSE loss $\mathcal{L}_{\text{MSE}}$ directly anchors predictions to known binding affinity labels. This complementary combination allows DualBind to capture both local structural variations and global binding trends effectively. An overview of our DualBind approach is illustrated in Figure~\ref{fig:DualBind}.

\textit{MSE Loss.} As shown in the upper branch of Figure~\ref{fig:DualBind}, the MSE loss serves as a direct supervision signal for affinity prediction. Given a protein-ligand complex structure $(\boldsymbol{A}, \boldsymbol{X})$, the model predicts its binding affinity, and the error is calculated against the ground truth value. Formally, for a single data sample,
\begin{equation}
    \label{eq:loss_mse}
    \mathcal{L}_{\text{MSE}} = \left(E_\theta(\boldsymbol{A}, \boldsymbol{X}) - y\right)^2,
\end{equation}
where $y$ represents the ground truth binding affinity, which corresponds to the AB-FEP-calculated label in ToxBench data. Intuitively, the MSE loss ensures that points, representing observed complex structures within the energy landscape, remain anchored to these affinity labels, preserving their alignment with the training data.

\textit{DSM Loss.} Training neural networks with denoising is a well-established technique for enhancing model robustness and generalization~\citep{bishop1995training,vincent2008extracting,kong2020flag,godwin2021simple}. In particular, the denoising score matching (DSM) technique has been employed to pretrain models for 3D molecular tasks, demonstrating that such objective not only simulates learning a molecular force field but also significantly boosts performance across various downstream tasks~\citep{zaidi2022pre}. Furthermore, within the framework of energy-based models (EBMs), the DSM objective can be interpreted as modeling the likelihood of observed data~\citep{song2019generative,ho2020denoising,song2021maximum,jin2023dsmbind}. By training an EBM to denoise perturbed data, we implicitly maximize the likelihood of observing the original, unperturbed data under the modeled distribution. Thus, in the context of protein-ligand binding energy modeling, the DSM objective ensures that \textit{the original complex structure is guided toward the local minima of the learned energy landscape}

Given such generalization capability and ability to shape local minima for observed complex structures, we incorporate the DSM objective into our DualBind model. Following previous DSM-based studies~\citep{zaidi2022pre, jin2023unsupervised}, given a complex struture $(\boldsymbol{A}, \boldsymbol{X})$ from the training dataset, we first perturb it by adding Gaussian noise specifically to the ligand atom coordinates. Formally, for each ligand atom coordinate $\boldsymbol{x}_i$, the perturbed coordinate $\boldsymbol{\tilde{x}}_i$ can be obtained by
\begin{equation}
    \label{eq:perturbation}
    \boldsymbol{\tilde{x}}_i = \boldsymbol{x}_i + \sigma\boldsymbol{\epsilon}_i, \quad \text{where } \boldsymbol{\epsilon}_i \sim \mathcal{N}(0, \boldsymbol{I}_3).
\end{equation}
$\sigma$ is a hyperparameter that controls the noise scale. Note that noise is added only to ligand atoms, while protein atoms remain fixed. For simplicity, we denote the full coordinate matrix, including perturbed ligand atoms and unperturbed protein atoms, as $\boldsymbol{\tilde{X}}$.

As illustrated in the lower branch in Figure~\ref{fig:DualBind}, the perturbed structure is fed into the model to obtain the predicted energy $E_\theta(\boldsymbol{A}, \boldsymbol{\tilde{X}})$. Then the DSM loss matches the score of the model distribution $p_\theta$ at the perturbed data $\boldsymbol{\tilde{X}}$ with the score of the distribution $q$ where $\boldsymbol{\tilde{X}}$ is sampled. Score is defined as the gradient of
the log-probability \emph{w.r.t.} $\boldsymbol{\tilde{X}}$. Formally,
\begin{equation}
    \label{eq:dsm_loss_initial}
    \mathcal{L}_{\text{DSM}} = \mathbb{E}_{q(\boldsymbol{\tilde{X}})} \left[\left\|\nabla_{\boldsymbol{\tilde{X}}}\log p_\theta(\boldsymbol{A}, \boldsymbol{\tilde{X}}) - \nabla_{\boldsymbol{\tilde{X}}}\log q(\boldsymbol{\tilde{X}})\right\|^2\right].
\end{equation}
As shown by~\citet{vincent2011connection}, this is equivalent to
\begin{equation}
    \label{eq:dsm_loss}
    \begin{aligned}
    \mathcal{L}_{\text{DSM}} = \mathbb{E}_{q(\boldsymbol{\tilde{X}}|\boldsymbol{X})p_{\text{data}}(\boldsymbol{X})} &\left[\left\|\nabla_{\boldsymbol{\tilde{X}}}\log p_\theta(\boldsymbol{A}, \boldsymbol{\tilde{X}})\right.\right. \\
    & \left.\left. - \nabla_{\boldsymbol{\tilde{X}}}\log q(\boldsymbol{\tilde{X}}|\boldsymbol{X})\right\|^2\right].
    \end{aligned}
\end{equation}
The model’s energy function defines the probability distribution as $p_\theta(\boldsymbol{A}, \boldsymbol{X})=\frac{\exp{(-E_\theta(\boldsymbol{A}, \boldsymbol{X}))}}{Z_\theta}$, where $Z_\theta$ is the normalizing constant, which is typically intractable but a constant \emph{w.r.t.} $\boldsymbol{X}$~\citep{lecun2006tutorial,song2021train}. Therefore, the first term in Eq.~(\ref{eq:dsm_loss}) can be computed by
\begin{equation}
    \label{eq:dsm_loss_first_term}
    \begin{aligned}
        \nabla_{\boldsymbol{\tilde{X}}}\log p_\theta(\boldsymbol{A}, \boldsymbol{\tilde{X}}) &= -\nabla_{\boldsymbol{\tilde{X}}}E_\theta(\boldsymbol{A}, \boldsymbol{\tilde{X}}) - \nabla_{\boldsymbol{\tilde{X}}}\log Z_\theta \\
        &= -\nabla_{\boldsymbol{\tilde{X}}}E_\theta(\boldsymbol{A}, \boldsymbol{\tilde{X}}).
    \end{aligned}
\end{equation}
The second term in Eq.~(\ref{eq:dsm_loss}) can be easily computed as $\nabla_{\boldsymbol{\tilde{X}}}\log q(\boldsymbol{\tilde{X}}|\boldsymbol{X})=-\frac{\boldsymbol{(\tilde{X}}-\boldsymbol{X})}{\sigma^2}$ since $q(\boldsymbol{\tilde{X}}|\boldsymbol{X})$ is a Gaussian distribution. Hence, 
\begin{equation}
    \label{eq:dsm_loss_final}
    \begin{aligned}
    \mathcal{L}_{\text{DSM}} = \mathbb{E}_{q(\boldsymbol{\tilde{X}}|\boldsymbol{X})p_{\text{data}}(\boldsymbol{X})} &\left[\left\|\nabla_{\boldsymbol{\tilde{X}}}E_\theta(\boldsymbol{A}, \boldsymbol{\tilde{X}}) - \frac{\boldsymbol{(\tilde{X}}-\boldsymbol{X})}{\sigma^2}\right\|^2\right].
    \end{aligned}
\end{equation}
Intuitively, minimizing this loss encourages the model to shape its energy landscape such that energy valleys (\emph{a.k.a.}, local minima) align with the original unperturbed protein-ligand structures.

The final training loss is defined as a weighted sum of the MSE loss and the DSM loss, \emph{i.e.},
\begin{equation}
    \label{eq:total_loss}
    \mathcal{L} = \mathcal{L}_{\text{MSE}} + \lambda\mathcal{L}_{\text{DSM}},
\end{equation}
where $\lambda$ is a hyperparameter that balances the contribution of the two losses.

\textbf{The Parameterization of $E_\theta(\boldsymbol{A}, \boldsymbol{X})$.} It is important to note that our dual-loss framework is model-agnostic, allowing seamless integration with various model architectures for $E_\theta(\boldsymbol{A}, \boldsymbol{X})$. In this work, we adopt the architecture from~\citet{jin2023unsupervised}, which employs an SE(3)-invariant model, built on a frame averaging neural network~\citep{puny2021frame}, to obtain atom representations. The key difference is that we use attention layers~\citep{vaswani2017attention} within the frame averaging framework rather than the SRU++ architecture~\citep{lei2021attention}. Then, binding affinity predictions are computed by capturing pairwise atom interactions within a predefined distance threshold. For further details on this architecture, we refer readers to~\citet{jin2023unsupervised}.

\section{Experiments}
\label{sec:experiments}

In this section, we describe the experimental setup, present the benchmarking results, and analyze the findings.

\textbf{Benchmarked Models}. In addition to our proposed DualBind, we include two representative baselines in our benchmark study. Recognizing that an exhaustive evaluation of all published ML models for binding affinity prediction is impractical, our selection aims to demonstrate the performance of distinct model designs on ToxBench and establish initial baselines to facilitate future research using this dataset. The included models are:
\begin{itemize}
    \item \textit{Chemprop}~\citep{heid2023chemprop}. We include Chemprop, a widely-adopted model for molecular property prediction, as a ligand-only baseline. It uses a directed message passing neural network~\citep{yang2019analyzing} that operates solely on the 2D molecular graph of the ligand to predict the binding affinity, without incorporating any information about the protein. Its inclusion allows us to establish the performance level achievable using only ligand information within the ToxBench context, providing an important reference point for evaluating interaction-aware models.
    \item \textit{AEV-PLIG}~\citep{warren2024make}. AEV-PLIG is a recently proposed interaction-aware model for binding affinity prediction. AEV-PLIG first uses atomic environment vectors (AEVs)~\citep{smith2017ani} to capture local atomic chemical environments derived from the 3D protein-ligand complex. These pre-computed AEV features are then processed by a graph attention network~\citep{brodyattentive} to predict binding affinity. This approach, based on fixed pre-computed structural features, differs from our DualBind model, which is designed to learn relevant geometric representations directly from 3D structures.
    \item \textit{DualBind}. As detailed in Section~\ref{sec:DualBind}, our proposed DualBind employs a 3D-invariant model trained with a novel dual-loss strategy to predict binding affinity from the 3D protein-ligand complex.
\end{itemize}

\textbf{Evaluation Metrics}. We use the following standard regression metrics to quantitatively evaluate performance on the ToxBench task.
\begin{itemize}
    \item \textit{Pearson Correlation Coefficient ($R_p$)}. This measures the linear correlation between the predicted affinities and the ground truth affinities. Values range from -1 to 1, where 1 denotes perfect positive linear correlation, 0 indicates no linear correlation, and -1 indicates perfect negative linear correlation. Higher positive values are better.
    \item \textit{Coefficient of Determination ($R^2$)}. Also known as R-squared, this metric represents the proportion of the variance in the ground truth affinities that is predictable from the predicted affinities. $R^2$ values closer to 1 indicate a better fit of the model to the data.
    \item \textit{Spearman Rank Correlation Coefficient ($\rho$)}: This metric assesses the strength and direction of the monotonic relationship between the ranks of the predicted and true affinities. Ranging from -1 to 1, it is less sensitive to outliers than Pearson $R_p$ and is evaluating a model's ability to correctly rank ligands according to their binding affinity, a critical aspect in virtual screening scenarios. Values closer to 1 indicate better ranking performance.
    \item \textit{Root Mean Square Error (RMSE)}. RMSE measures the standard deviation of the prediction errors, providing a measure of the absolute error magnitude in the units of the target variable (\emph{i.e.}, \textit{kcal/mol} in the ToxBench context). Lower RMSE values indicate better predictive accuracy.
\end{itemize}

\textbf{Implementation Details}. We train and evaluate all three models using the predefined ToxBench splits. Final models for test set reporting are selected based on the checkpoint achieving the lowest RMSE on the validation set. Acknowledging that AB-FEP-calculated affinities greater than -3.0~\textit{kcal/mol} typically indicate non-binding or very weak interactions, we apply a thresholding approach in our experiments. Specifically, ToxBench labels with original affinity values exceeding -3.0 \textit{kcal/mol} are capped at -3.0~\textit{kcal/mol} prior to training. Correspondingly, during inference, predicted affinity values greater than -3.0~\textit{kcal/mol} are also adjusted to -3.0~\textit{kcal/mol} before performance metrics are computed. For robustness, all models are trained with three independent random seeds and the average performance metrics are reported. 

\begin{figure*}[t]
	\centering
\includegraphics[width=\textwidth]{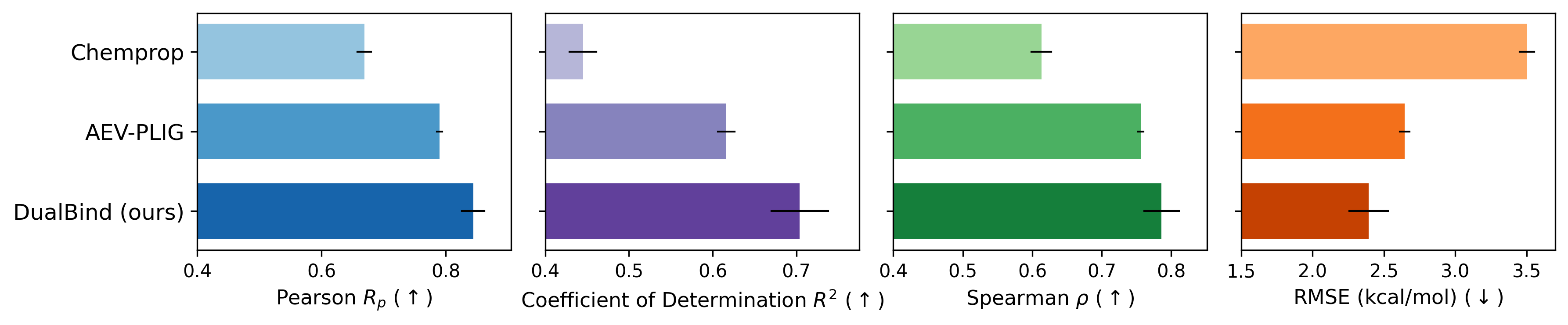}
	\caption{\textbf{Performance comparison on the ToxBench benchmark.} The evaluation metrics include $R_p$ (Pearson correlation coefficient), $R^2$ (coefficient of determination), $\rho$ (Spearman's rank correlation coefficient), and RMSE (root mean square error). $\uparrow$ ($\downarrow$) represents that a higher (lower) value denotes better performance. Results presented are the averages obtained from three independent runs for each method. Detailed numerical results are provided in Table~\ref{tab:toxbench_perf}.}
	\label{fig:toxbench_perf}
\end{figure*}

For Chemprop and AEV-PLIG, we adapt the official open-sourced implementations and train them on the ToxBench dataset. For ChemProp, we use the default parameters in their v2.1.0 release. For AEV-PLIG, we use the optimized hyperparameters reported in their official implementations as a starting point. Key hyperparameters, such as learning rate and training epochs, are subsequently fine-tuned by monitoring the RMSE on the ToxBench validation set.

Our implementation of DualBind is based on PyTorch~\citep{paszke2017automatic} and PyTorch Lightning~\citep{falcon2019pytorch}. For protein structure input, DualBind uses a local 3D region comprising the 50 residues closest to the ligand. During training, the noise scale $\sigma$ for the DSM objective, as defined in Eq.~(\ref{eq:perturbation}), is uniformly sampled from the interval [0.1, 1]. The weight $\lambda$ for the DSM loss component is set to 2. We employ the Adam optimizer~\citep{kingma2014adam} as our training optimizer. Key hyperparameters, including learning rate, batch size, and dropout rate, are tuned against the ToxBench validation set. The final configuration uses a learning rate of $5\times 10^{-4}$, a batch size of 128, and a dropout rate of 0.1. The learning rate is decayed by a factor of 0.95 after each epoch. The model is trained for a total of 120 epochs. Training for the DualBind model is conducted on 8 NVIDIA A100 GPUs and typically completed in $\sim$4 hours. The model has a total of $\sim$1.02 million learnable parameters.

\begin{figure}[t]
	\centering
\includegraphics[width=0.3\textwidth]{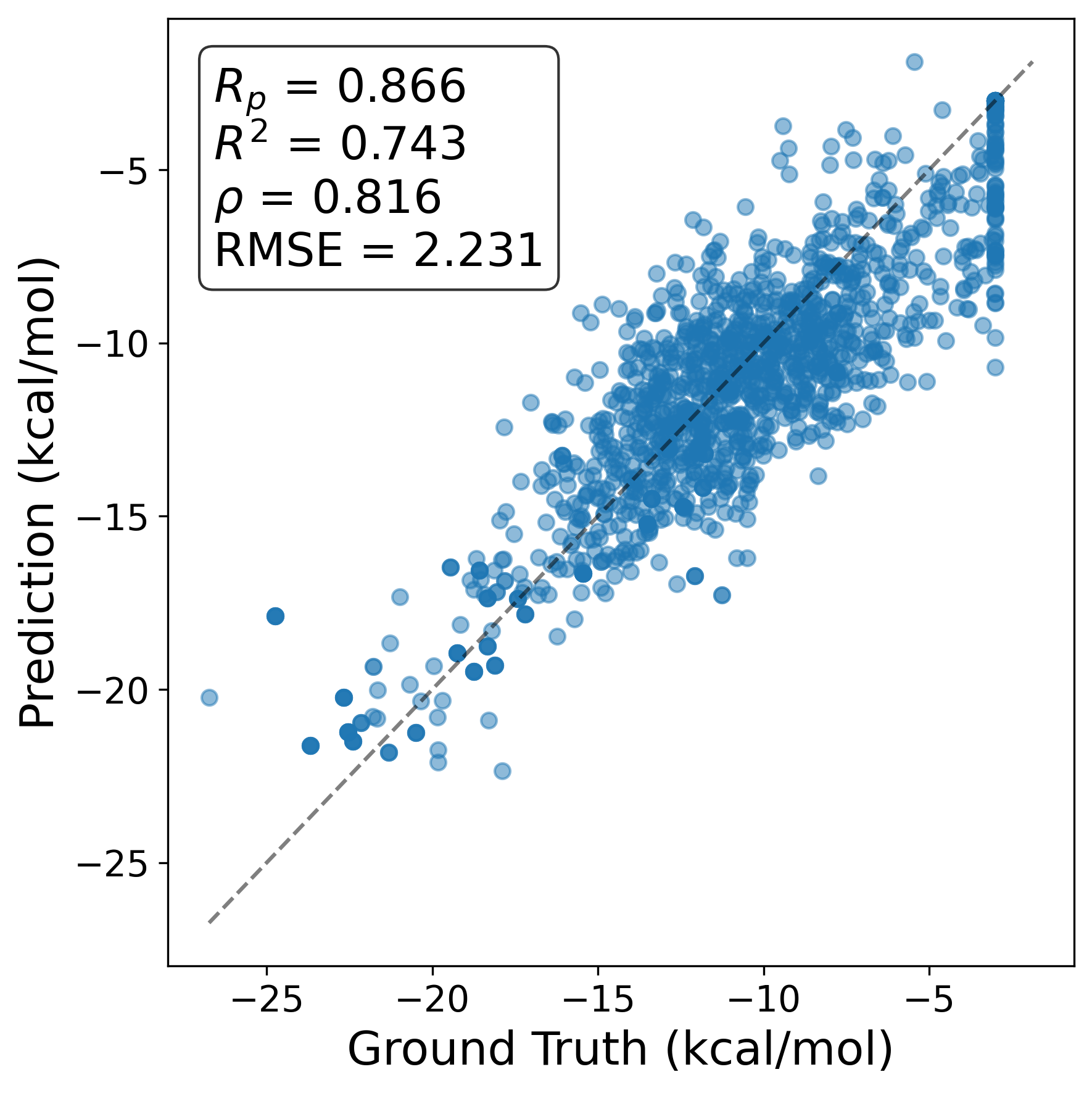}
	\caption{\textbf{DualBind predictions \emph{vs.} ground truth}. This compares DualBind's predicted binding affinities from the best performing of three runs against the corresponding ground truth values on the ToxBench test set.}
	\label{fig:dualbind_pred_scatter}
\end{figure}

\textbf{Benchmarking Results}. We evaluate the trained Chemprop, AEV-PLIG, and our proposed DualBind model on the ToxBench test set using the previously defined metrics. Figure~\ref{fig:toxbench_perf} visually summarizes the comparative performance, while the precise numerical results are detailed in Table~\ref{tab:toxbench_perf}. Our proposed DualBind model consistently outperforms both Chemprop and AEV-PLIG across all evaluation metrics. DualBind achieves the highest Pearson $R_p$ of 0.844, the highest $R^2$ of 0.704, the highest Spearman $\rho$ of 0.786, and the lowest RMSE of 2.392~\textit{kcal/mol}. Figure~\ref{fig:dualbind_pred_scatter} further provides a qualitative illustration of DualBind's strong predictive performance, presenting a scatter plot of affinities predicted by a single model (without ensembling) against the ground-truth AB-FEP-calculated values on the test set.

A particularly significant finding from our benchmark is the substantial performance gap observed between the ligand-only Chemprop model and the interaction-aware models, including DualBind and AEV-PLIG. Chemprop achieves an RMSE of 3.502 \textit{kcal/mol}, markedly worse than those achieved by AEV-PLIG (2.645~\textit{kcal/mol}) and DualBind (2.392~\textit{kcal/mol}). This advantage for interaction-aware models is also clear across all other metrics, as shown in Table~\ref{tab:toxbench_perf}. Such observed gap is critical when contextualized with the limitations of existing sparse, multi-target datasets like PDBBind. As discussed in Section~\ref{sec:related_work}, models trained on such benchmarks can achieve surprisingly strong performance by exploiting ligand-only or protein-only features due to inherent dataset biases, without necessarily learning true protein-ligand interactions~\citep{volkov2022frustration}. In contrast, the comparatively worse performance of the ligand-only model on ToxBench indicates that our dataset design, focusing on a single target with dense AB-FEP labeling, effectively mitigates such dataset-specific shortcuts. This design forces ML models to engage more deeply with the complexities of protein-ligand binding interactions to achieve high accuracy. In summary, the clear advantage demonstrated by interaction-aware models on ToxBench validates its utility as a robust benchmark for fostering the development of ML methods that truly learn the principles of molecular interaction for a specific target, a crucial step towards practical, target-centric drug discovery.

\begin{table}[t]
	\caption{\textbf{Performance comparison on the ToxBench benchmark.} Reported values for each method are the mean and standard deviation over three independent runs.}
	\label{tab:toxbench_perf}
	\centering
	\resizebox{\columnwidth}{!}{
	\begin{tabular}{lccccc}
		\toprule
		Method & $R_p$($\uparrow$) & $R^2$($\uparrow$) & $\rho$($\uparrow$) & RMSE($\downarrow$) \\
    \midrule
    Chemprop & $0.669\scriptstyle{{\light{\pm0.011}}}$ & $0.445\scriptstyle{{\light{\pm0.016}}}$ & $0.613\scriptstyle{{\light{\pm0.014}}}$ & $3.502\scriptstyle{{\light{\pm0.051}}}$ \\
    AEV-PLIG & $0.790\scriptstyle{{\light{\pm0.004}}}$ & $0.616\scriptstyle{{\light{\pm0.010}}}$ & $0.756\scriptstyle{{\light{\pm0.004}}}$ & $2.645\scriptstyle{{\light{\pm0.033}}}$ \\
    DualBind & $\textbf{0.844}\scriptstyle{{\light{\pm0.018}}}$ & $\textbf{0.704}\scriptstyle{{\light{\pm0.034}}}$ & $\textbf{0.786}\scriptstyle{{\light{\pm0.025}}}$ & $\textbf{2.392}\scriptstyle{{\light{\pm0.136}}}$ \\
    \bottomrule
	\end{tabular}
	}
\end{table}

\textbf{High-Throughput Affinity Prediction with ML}. In addition to assessing predictive accuracy, ToxBench is designed to unlock the potential of ML models to approximate binding affinities from high-fidelity physics-based methods like AB-FEP, but with a substantial reduction in computational cost. To be specific, a typical AB-FEP calculation for a single protein-ligand complex, such as those used to generate the labels in ToxBench, requires approximately 35 hours on an NVIDIA T4 GPU. In contrast, inference with the trained DualBind model is orders of magnitude faster. Measured over the ToxBench test set, DualBind's average inference time for a single complex on an NVIDIA A100 GPU is merely 126~\textit{ms} without batching. This further improves to an average of 33~\textit{ms} per complex when employing a practical batch size of 96. Remarkably, DualBind achieves a $10^6$-fold speed-up compared to the typical AB-FEP calculation time. Such a dramatic increase in throughput, coupled with the competitive predictive performance as described above (RMSE of 2.392~\textit{kcal/mol}), demonstrates the potential capability of carefully designed ML models like DualBind to approximate AB-FEP quality results at a fraction of the computational cost. This is significant for high-throughput binding affinity prediction, allowing for rapid screening of large-scale chemical libraries to accelerate target-based drug discovery.


\section{Conclusion and Outlook}

In this work, we introduce ToxBench, a novel benchmark dataset comprising AB-FEP-calculated binding affinities specifically for Human Estrogen Receptor Alpha, a pharmaceutically critical target. ToxBench is designed with a single-target focus to facilitate the development of ML models that genuinely learn protein-ligand interactions. Our benchmarking study validates this design, demonstrating that the ligand-only model significantly underperforms interaction-aware models on ToxBench, in contrast to the observations on prior benchmarks. Furthermore, our proposed DualBind model achieves state-of-the-art performance on this new benchmark. Importantly, our experiments show that ML models have great potential to approximate the high fidelity AB-FEP calculations while offering orders of magnitude computational speed-up, thereby paving the way for high-throughput applications.

In summary, ToxBench represents a significant advance towards more robust and practically relevant ML-driven binding affinity prediction. It offers a valuable testbed for the community to develop, evaluate, and compare ML models. Moreover, the design principles behind ToxBench provide a blueprint for creating similarly impactful datasets across other key biological targets.

\nocite{langley00}

\bibliography{references}
\bibliographystyle{icml2025}



\end{document}